\def\year{2021}\relax
\documentclass[letterpaper]{article} % DO NOT CHANGE THIS
\usepackage{booktabs}
\usepackage[T1]{fontenc}
\usepackage{aaai21}  % DO NOT CHANGE THIS
\usepackage{times}  % DO NOT CHANGE THIS
\usepackage{helvet} % DO NOT CHANGE THIS
\usepackage{courier}  % DO NOT CHANGE THIS
\usepackage[hyphens]{url}  % DO NOT CHANGE THIS
\usepackage{graphicx} % DO NOT CHANGE THIS
\usepackage{natbib}  % DO NOT CHANGE THIS AND DO NOT ADD ANY OPTIONS TO IT
\usepackage{caption} % DO NOT CHANGE THIS AND DO NOT ADD ANY OPTIONS TO IT
\usepackage{amsmath}
\usepackage{amsmath} 
\newcommand{\bftab}{\fontseries{b}\selectfont}

\title{Factor Graph Molecule Network for Structure Elucidation}
\author{
    Le Trung Hieu, \textsuperscript{\rm 1}
    Xu Yiqing, \textsuperscript{\rm 1}
    Lee Wee Sun, \textsuperscript{\rm 1}
}
\affiliations{
    \textsuperscript{\rm 1} National University of Singapore \\
    hieu.le@comp.nus.edu.sg, xu.yiqing@comp.nus.edu.sg, leews@comp.nus.edu.sg	
    
}
\begin{document}

\maketitle

\begin{abstract}
Designing a network to learn a molecule structure given its physical/chemical properties is a hard problem, but is useful for drug discovery tasks. In this paper, we incorporate higher-order relational learning of Factor Graphs with strong approximation power of Neural Networks to create a molecule-structure learning network that has strong generalization power and can enforce higher-order relationship and valence constraints. We further propose methods to tackle problems such as the efficient design of factor nodes, conditional parameter sharing among factors, and symmetry problems in molecule structure prediction. 
%To our knowledge, this is the first work on molecule structure prediction that does not require third-party software to generate possible substructures and rank them, making our model inference process less dependent on domain knowledge. 
Our experiment evaluation shows that the factor learning is effective and outperforms related methods.
\end{abstract}

\section{Introduction}
% \begin{itemize}
%     \item Machine learning has been useful for the task of drug discovery. 1-2 more sentences explain why drug discovery is important. 
%     \item Drug discovery has been divided into two separate areas: molecule properties prediction given structure, and molecule structure predictions given measured properties.
%     \item Many works have been done for the first area, whereas the second area is more difficult to solve (search space is too huge and no efficient way to enforce constraint yet).
%     \item Short intro to our proposed architecture, and why is it useful.
% \end{itemize}

The identification of a Chemistry Molecule Structure given its measured physical/chemical properties is an important task for drug discovery. Discovering the structure will aid scientists to have better information about how the drugs will behave in various scenarios. 

\begin{figure}[!tbp]
  \centering
  \begin{minipage}[b]{0.4\textwidth}
    \includegraphics[width=\textwidth]{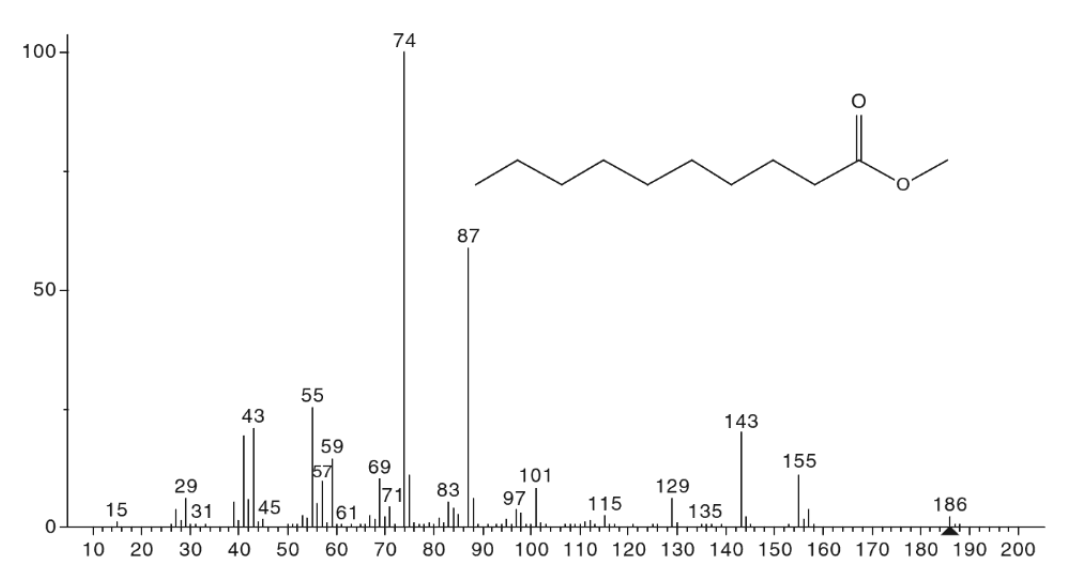}
    \caption{Mass Spectrum}
    \cite{kitson1996gas}
    \label{mass_spectrum}
  \end{minipage}
  \hfill
\end{figure}

In this paper, we will elucidate chemistry structure by leveraging the measurement of the molecule mass spectrum, which is a plot of the ion signal as
a function of the mass-to-charge ratio, an indication of detected ions abundance. Traditional methods require a chemist to match components of spectrum against a library of known spectra, then subsequently do result combination. Such process is tedious and time-consuming. Figure \ref{mass_spectrum} contains an example mass spectrum and the corresponding molecule structures. The task is, given the mass spectrum and the molecule formula (e.g C$_4$H$_5$O$_6$), we need to predict the chemical structure of the compound.

Works have been done for the above task using traditional Machine Learning methods \cite{lim2018chemical}, but results are still not satisfactory. %A natural question prompts for better methods to solve this task.
Deep Learning has led to breakthroughs in Computer Vision \cite{krizhevsky2012imagenet} and Natural Language Processing \cite{vaswani2017attention}, so it is natural to ask whether it would perform well for other areas such as Drug Discovery. However, many of the promising works in this field \cite{maziarka2020molecule} mainly focus on predicting the molecule properties given its structure, instead of structure elucidation given some observed measurements. Indeed, latter task is much harder than the former, because of the large search space given the large number of potential atom edge combinations, and the lack of effective learning methods to capture the higher-order relationship between the nodes as well as the valence constraints.

To tackle the challenge, in this paper we propose the Molecule Factor Graph Network (MFGN), which incorporates Factor Graph into a Neural Network  to better leverage the higher-order relationship learning between the nodes. We adapt the Low Rank Belief Propagation Neural Network \cite{dupty2020neuralizing} to the chemical structures prediction task by proposing various effective design for different types of graph nodes and factor nodes tailored for the task. These architecture elements capture the higher-order relationships and enforces hard constraints such as the valence constraint of each atom. We also remove the need for substructure generated by external softwares, relying mainly on a learning approach. %which makes our approach becomes free of domain-knowledge, which existing works lack of.
The experiment result shows performance improvement using the proposed model. We also conduct various ablation study and analysis to understand how each of our components behave.

\section{Related Work}
\subsection{Chemical Structure Elucidation (CSE)}
One notable work on molecular structure prediction is computer-aided CSE from mass spectrometry (MS) by matching substructures \cite{lim2018chemical}, because substructures might be responsible for certain m/z peaks in a mass spectra. In early work \cite{varmuza1996mass}, the authors developed a set of mass spectral classifiers to recognize the presence or absence of substructures or more general structural properties in a molecule, then they used the predicted substructures to reduce the number of candidates. Later work \cite{kerber2001molgen}\cite{schymanski2008use}\cite{schymanski2011automated} combined a software named MOLGEN-MS \cite{kerber2001molgen} with additional structural information and made the number of candidates much smaller. The latest state-of-the-art work \cite{lim2018chemical} combines the substructure classification results to the final list of candidate structures through graph isomorphism methods.

Such approaches require domain knowledge to generate the possible substructures to feed the classifiers, usually through third-party softwares. Our paper contributes a new approach that eliminates the need to rely on external software for the task.

\subsection{Deep Learning for CSE}

Deep Learning has been successfully applied to the field of Chemistry and Drug Discovery, but limited to molecule property prediction task given its known structures. Two current state-of-the-art methods in this field are variants of Message Passing Neural Network (MPNN) \cite{gilmer2017neural} and Transformer \cite{maziarka2020molecule}, in which their neural architectures allow efficient interaction between atoms based on the bond and distance of pairwise edge connection. We will extend the MPNN and Transformer application above to our task, as the baselines. 

In general, the extension of deep learning for structure prediction is difficult \cite{lim2018chemical} with current state-of-the-art models that use Graph Neural Network (GNN) style architecture. The reason is that, GNNs mostly capture first order approximations in aggregating information from the neighbouring nodes, but fails to capture the more complex higher-order constraints in a molecule structure prediction task (e.g valence constraints of each atom, or the joint-information from each mass spectrum peak to a set of other atoms that represent a likely substructure)

\subsection{Factor Graph Neuralization}

Node interactions have also been modeled with Probabilistic Graphical Models (PGM). A common graphical representation of PGM that is able to captures higher-order constraints of nodes is the Factor Graph, which include factor nodes that represents a higher-order relationship between a subset of graph nodes. There is rich background of theoretically-grounded PGM algorithms to infer a state of a node, given its statistical relation with the factors. Knowledge of these inference algorithm produce good inductive bias that can lead to better generalization of neural networks, if incorporated correctly \cite{battaglia2018relational}.

A successful attempt at Factor Graph Neuralization comes from \cite{dupty2020neuralizing}, where the factors are approximated using tensor decomposition, and in turn can be abstracted into a neural network module that can be learned end-to-end within a bigger architecture. We will cover more details in Preliminary section. Inspired by their idea, we will design a better network that enforce the necessary special higher-order constraints in CSE task.

\section{Preliminary}
\subsection{Basic Chemistry}
We will briefly introduce about two important chemistry concepts here. Mass spectrum requires a soft higher-order relationship between each spectrum peak to all of the molecule atoms, while valence constraint enforce a hard high-order constraint between the sum of edges bonds for each atom. These two higher-order relationships will be captured in our architecture.

\begin{itemize}
    \item \textbf{Mass spectrum:} The mass spectrum represents the distribution of ions by mass in a molecule to reflect the relative abundance of detected ions (Figure \ref{mass_spectrum}). The x-axis is the mass-to-charge scale (that represents a relationship between the mass of a given ion and the number of elementary charges that it carries), and the y-axis is the signal intensity scale of the ions. The mass spectrum will contain peaks that represent fragment ions as well as the molecular ion \cite{kitson1996gas}. In Figure \ref{mass_spectrum}, the x-axis value of 186 represents the molecule mass of decanoic acid substructure C$_{11}$H$_{22}$O$_2$ (atom mass sum equal 186), but if we analyze it manually, it could be some other fragments with the same total mass like C$_{10}$H$_{18}$O$_3$. Therefore, it is important for the network to capture the higher-order relationship between each mass spectrum peak to all the atoms of the molecules.
    \item \textbf{Valence Constraint:} From the structural view, a molecule consists of the aggregation of different atoms held together by valence forces, hence the structure must also obey the valence rules. For example, the valence of Carbon is 4 and valence of H is 1, therefore 1 Carbon cannot connect with more than 4 Hs. So the sum of edge bonds connecting to each atom must be equal to its valence, and enforcing this hard constraint is difficult . Inspired from error correcting code, we will design a factor node enforcing valence constraint in our network as explained later.
\end{itemize}

\subsection{Factor Graph Neuralization}

\subsubsection{Graphical Models and Loopy Belief Propagation (LBP)}

Consider a factor graph $G = (V, E)$ where vertices $V = {1, 2, ..., n}$ and hyperedges $E \in 2^V$. Each hyperedge represent a factor node that connect to a subset of vertices. Every vertex $v_i \in V$ is associated with discrete random variable $x_i$, while every hyperedge $e \in E$ is associated with a potential function (of a factor node $a$) $f_a \in F$. In this paper, we will index vertices as integers ${0, 1, 2, ..., i, j}$ and factor nodes as ${a, b, c, ...}$

Loopy Belief Propagation (LBP) is the process of computing and passing messages between factor nodes and vertices to compute approximate marginal probability $p(x_i)$ for $x_i \in X$. LBP initializes two kinds of messages, factor-to-node $m_{a \rightarrow i}(x_i)$ and node-to-factor $m_{i \rightarrow a}(x_i)$. The recursive update rule for the message is as follows:

\begin{equation}
    m_{i \rightarrow a}(x_i) = \prod_{c \in N(i) \setminus a} m_{c \rightarrow i}(x_i)
\end{equation}

\begin{equation}
    m_{a \rightarrow i}(x_i) = \sum_{X_a \setminus {x_i}} f_a(X_a) \prod_{j \in N(a) \setminus {i}} m_{j \rightarrow a}(x_j)
\end{equation}

whereas $N(i)$ is the set of neighbours of $i$. After sufficient number of recursive iterations, the belief of variable is computed as :

\begin{equation}
    b_i(x_i) = f_i(x_i) \prod_{a \in N(i)} m_{a \rightarrow i}(x_i)
\end{equation}

\subsubsection{Low-rank LBP}
In \cite{dupty2020neuralizing}, each factor is approximated as a low-rank tensor. By applying low-rank tensor decomposition, the message update rule becomes:
    % m_{a \rightarrow i} = W_{a_i}[(W_{a_1}^T m_{1 \rightarrow a}) \odot (W_{a_2}^T m_{2 \rightarrow a}) \odot (W_{a_n}^T m_{n \rightarrow a}) ]
    
\begin{equation}
    m_{a \rightarrow i} = 
    W_{a_i} [\bigodot_{j \in N(a) \setminus{i}} (W_{a_j}^T m_{j \rightarrow a}) ]
    % W_{a_i} [\bigodot_{j \in N(a) \setminus{i}} (W_{a_j}^T m_{j \rightarrow a})]
\end{equation}

\begin{equation}
    m_{i \rightarrow a} = \bigodot_{c \in N(i) \setminus {a}} m_{c \rightarrow i}
\end{equation}
where $\bigodot$ denotes component-wise multiplication.

With this derivation, the computational complexity of messages grow linearly with addition of number of variables to factors. Therefore, this inference algorithm is efficient enough to be used in an end-to-end neural network.

\subsubsection{Neuralizing low-rank LBP}

\begin{figure}[!tbp]
  \centering
  \begin{minipage}[b]{0.4\textwidth}
    \includegraphics[width=\textwidth]{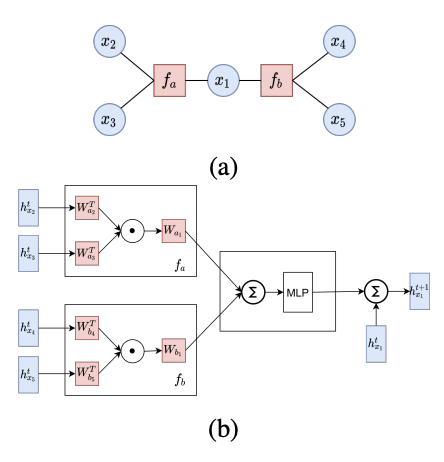}
    \caption{(a) Factor graph with $x_1$ and its
neighbours. (b) Message passing update
for node $x_1$\cite{dupty2020neuralizing}}
    \label{factor_graph_neuralize}
  \end{minipage}
  \hfill
\end{figure}

With the efficient derivations in (4) and (5), we can derive a neuralization that incorporates structural inductive bias to the neural network using the factor graph inference learning \cite{dupty2020neuralizing}. The LBP algorithm neuralization in (4) and (5) contains chained multiplication of numerous terms leading to numerical instability. Empirically, \citet{dupty2020neuralizing} found that replacing the multiplication operation in the message update function with a sum operation, followed by a MLP works well in learning useful representations. Let graph $G = (V, E)$ with nodes and edges, and $F$ be the factor graph defined on top of $G$. Let $h_{x_i}^t$ be hidden state of node $x_i$ at iteration t. Then, the network will have the following message passing update: 

\begin{equation} \label{neuralize}
    h_{x_i}^{t+1} = h_{x_i}^t + MLP(\sum_{a \in N(i)} W_{a_i} [\bigodot_{j \in N(a) \setminus {i}} W_{a_j}^T h_{x_j}^t])
\end{equation}

Here we are using node embedding $h_{x_i}$ as approximation for message embedding, which is simpler and works well empirically \cite{dupty2020neuralizing}. The component of this message update neuralization is depicted in Figure \ref{factor_graph_neuralize}

\section{Molecule Factor Graph Network}
\begin{figure*}[!tbp]
  \centering
  \begin{minipage}[b]{1\textwidth}
    \includegraphics[width=\textwidth]{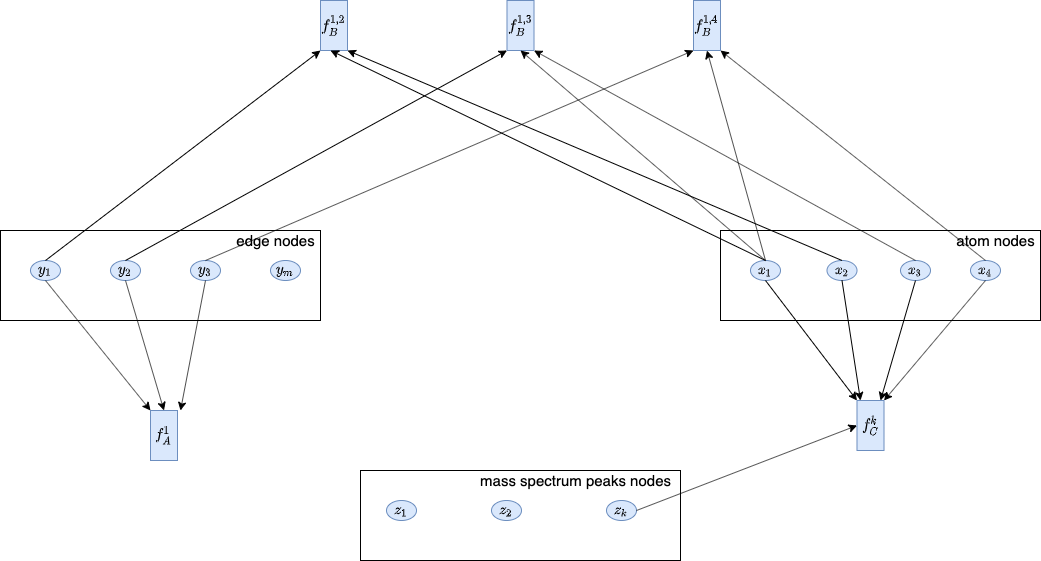}
    \caption{FGMN Architecture}
    \label{factor_graph_neuralize}
  \end{minipage}
  \hfill
\end{figure*}

In this section, we will describe our proposed model architecture together with the pitfalls and design considerations. Following \cite{dupty2020neuralizing} (our Preliminary Section), we will design a Graph Neural Network with different types of factors. There will be 3 types of graph vertices here: atom nodes, edge nodes, mass spectrum nodes.
We first recap the problem formulation: Given k mass spectrum nodes representing the peaks $z_1, z_2, \dots, z_k$, predict the adjacency matrix $\mathcal{A}$ representing the molecule structures, where vertices in the matrix are list of $n$ atom nodes $x_1, x_2, \dots, x_n$, and entries in the matrix are list of $m$ edge nodes $y_1, y_2, \dots, y_m$.

With these three types of graph nodes, we need to define the factors that capture the higher-order relationship among them. In Preliminary section (Basic Chemistry), we mentioned two most important constraint here, first is that a mass spectrum peak indicates the presence of a certain substructure, and the second is the valence constraint of sum of edges bonds connecting to the same atom.

\subsection{Graph Factors}
\begin{itemize}
    \item \textbf{MSP-Atom factor (Type C)}: The $k_{th}$ factor C $f_C^k$ will connect with mass spectrum peak node $z_k$ and all the $n$ atoms of this current molecule $x_1, x_2, \dots, x_n$. This factor node design is based on the fact that each peak index represents the total mass of a substructure fragment. Therefore, when connecting a mass spectrum node to all the molecule atom nodes (atom nodes embed the atom mass in its value as well), more information will flow from this mass spectrum peak to the atoms in corresponding substructures during message update, which aids the learning and inference.
    \item \textbf{Edge-Atom factor (Type B)}: Each edge node $y_l$ and the two atoms linked by the edge are connected to a factor type B $f_B^{i, j}$. This factor is responsible for the relationship of each edge and its two atoms.
    \item \textbf{Edge-Edge factor (Type A)}: Each atom will holds a valence constraint (e.g Carbon valence is 4, Oxygen valence is 2). For each atom $x_i$, we find all $t$ edges $y_1, y_2, \dots, y_t$ which have atom $x_i$ at one of two ends of the edge, then factor type A $f_A^i$ will connect to all these $t$ edges, and enforce the sum of these edges value to equal the valence value of the atom $x_i$. 
\end{itemize}

\subsection{Graph Factors Implementation}

\begin{itemize}
    \item \textbf{Type B and Type C}: The two factor types here connect to a low number of nodes. In this work, we will approximate these factors as low-rank tensors and adapt derivations from Equation (\ref{neuralize}).
    \item \textbf{Type A}: This is a hard-constraint factor node that is enforcing the sum of the nodes connected to a certain value. We will derive as following in the following section.
\end{itemize}

\subsubsection{Valence Factor Type A Formula}

We will derive the message from factor to each variable through a Dynamic Programming formula. The idea is inspired by Parity Check in Error-Correcting code formula derivation \cite{mncN}, in which each of its factor tries to enforce the XOR operations of all nodes connected to zero.

Let $\mu^{valence}_{m}$ be the message from a valence factor type A to a specific atom variable with valence constraint equal to $valence$, using only the first $m$ variables connected to the factor ($ m \in {1, 2, \dots}$. Let $g_m(x)$ be the message from variable $m$ to the factor, with value of variable as $x$. Denote $b$ as maximum value of $x$. Then the message from factor to variable is equivalent to the following derivation (Note that $g_i(x)$ is the message from variable $i$ to this current factor):

\begin{equation}
\begin{split} 
 \mu^{v}_{m} & = \sum_{{x_i}: \sum_{i=1}^m x_i = v}\prod_{i=1}^{m}(g_i(x_i)) \\
& = g_m(0) * \sum_{{x_i}: \sum_{i=1}^{m-1} x_i = v}\prod_{i=1}^{m-1}(g_i(x_i)) \\
&\qquad + g_m(1) * \sum_{{x_i}:\sum_{i=1}^{m-1} x_i = v-1}\prod_{i=1}^{m-1}(g_i(x_i)) \\
&\qquad   + \dots \\
&\qquad   + g_m(b) * \sum_{{x_i}:\sum_{i=1}^{m-1} x_i = v-b}\prod_{i=1}^{m-1}(g_i(x_i)) \\
& = g_m(0) * \mu^{v}_{m-1} + g_m(1) * \mu^{v-1}_{m-1} + ... + g_m(b) * \mu^{v-b}_{m-1}
\end{split}
\end{equation}

Therefore, from a brute force formula that is hard-to-implement and computationally expensive $\mathcal{O}(max\_bond^{max\_atoms})$, we have reduced to a Dynamic Programming formula efficient in performance and implementation with time complexity $\mathcal{O}(max\_bond\_type * max\_atoms * max\_value)$

\begin{equation}
\begin{split} 
    \mu^{v}_{m} = & g_m(0) * \mu^{v}_{m-1} + g_m(1) * \mu^{v-1}_{m-1} \\
    & + \dots + g_m(b) * \mu^{v-b}_{m-1}
\end{split}
\end{equation}

We also observe that this Dynamic Programming (DP) formula will contain a long chain of multiplication of the terms leading to numerical instability due to overflow and underflow issues. Because normalization does not alter final beliefs in Loopy Belief Propagation \cite{pearl2014probabilistic}, we are normalizing the hidden state of $g$ at each DP iteration.

\begin{table*}[hbt!]
    \label{test_result}
    \centering
    \caption{Results performance of each methods}
    \begin{tabular}{|p{5.8cm}|p{2cm}|p{2cm}|p{2cm}|p{2cm}|}
    \hline
    \multicolumn{1}{|c|}{ {Model} }& \multicolumn{2}{c|}{Train} &
    \multicolumn{2}{c|}{Test}\\
    \cline{2-5}
    \multicolumn{1}{|c|}{}&loss&acc&loss&acc\\
     \hline
    MASE with Encoder (non-positional encoding) & 0 & 1 & 0.403 & 0.8823 \\
    \hline
    MASE with Encoder (positional encoding) & 0 & 1 & 0.53 & 0.8901 \\
    \hline
    MASE with Encoder \& Decoder & 1.212 & 0.787 & 0.86 & 0.804 \\
    \hline
    MMPNN & 0.278 & 0.908 & 0.713 & 0.841 \\
    \hline
    FGMN-Low & 0.166 & 0.932 & 0.481 & 0.877 \\
    \hline
    \bftab FGMN-Medium & 0.059 &  0.989 & \bftab 0.275 & \bftab 0.914 \\
    \hline
    FGMN-High & 0.202 & 0.928 & 0.563 & 0.871 \\
    \hline
    \end{tabular}
\end{table*}

\subsection{Factor weights conditional sharing}

The factor weights are shared by conditioning on the nodes connected to the factor. We will define amount of shareable parameters as low, medium and high for each factor type B and type C (type A has no learnable parameter):

\subsubsection{Factor Type B Parameter Sharing}
\begin{itemize}
    \item \underline{Low}: Regardless of nodes connected, set of factor weights remains the same.
    \item \underline{Medium}: Factor weight set is conditioned on the atom type of two atoms connected to the factor, therefore number of different factor weight sets is $\text{max\_num\_atom\_types}^2$, \item \underline{High}: Factor weight set is conditioned on the atom index in the molecule, therefore number of different factor weight sets is $\text{max\_num\_atom\_per\_molecule}^2$
    % \item \textbf{MSP-Atom Type C}: \underline{Low}: $1$, \underline{Medium}: We will do a K-clustering on all the peaks we have from training dataset, then select top k peaks. Then for each factor \underline{High}
\end{itemize}

\begin{table*}[hbt!]
\begin{center}
\caption{Valence Satisfaction after Valence Factor Decoding Algorithm (Factor Type A), performed on various noise levels for ground truth}
 \begin{tabular}{||c c c c ||} 
 \hline
 Beta Noise Level & Initial  & Sum Factor & Multiply Factor\\ [0.5ex]
 \hline\hline
 0.1 & (Acc:1.000,valence:0.995) & (Acc:1.000,valence:0.997) & (Acc:1.000,valence:1.000) \\ 
 \hline
  0.2 & (Acc:0.991,valence:0.881) & (Acc:0.981,valence:0.768) & (Acc:0.987,valence:0.847) \\ 
 \hline
 0.5 & (Acc:0.811,valence:0.073) & (Acc:0.889,valence:0.267) & (Acc:0.893,valence:0.318) \\ 
 \hline
 1 & (Acc:0.565,valence:0) & (Acc:0.743,valence:0.128) & (Acc:0.775,valence:0.168) \\ 
 \hline
  2 & (Acc:0.398,valence:0.001) & (Acc:0.580,valence:0.027) &
  (Acc:0.650,valence:0.090) \\ 
 \hline
\end{tabular}
\end{center}
\end{table*}
\subsubsection{Factor Type C Parameter Sharing}
\begin{itemize}
    \item \underline{Low}: Regardless of nodes connected, set of factor weights remains the same.
    \item \underline{Medium}: We will do a $K$-means clustering on all the peaks we have from training dataset, then select top $k$ peaks. Then for each factor, the set of parameter is chosen based on the closest center from the clustering to the current mass spectrum node connected. Therefore, number of sharable factor weight set is $K$.
    \item \underline{High}: Dependent on the mass spectrum node value, which has around 1000 distinct values in our case.
\end{itemize}

\subsection{Symmetry Problem}
It is not easy to distinguish between first and subsequent atoms of the same type in the same molecule. First carbon connecting to first oxygen in the list of molecule atoms, or second carbon connecting to second oxygen in the list might actually represent the same molecule structure. Therefore, given the same inputs configurations, the training dataset might have different label structure while in fact just permutation matrix of each other. So, we need to devise good ordering strategy to approximate a canonical representation for isomorphic label structure matrices for during training in order to make learning easier.

We adopt the following ordering ways for the input data and label data. For the input data of edges, we use the atom mass based ordering, e.g. with lighter atom C at the front and heavier O at the back. Take the molecule Methyl decanoate C$_{11}$H$_{22}$O$_2$ for instance. The atoms list will be [C, C, C, C,C,C,C,C,C,C,C,O,O] and the input edges will be constructed by each pair of atoms.
For the label data of edges, we make use of the SMILES ordering since during training we know the molecular SMILES representation, which contains the information of molecular graph. The SMILES representation of this molecule is CCCCCCCCCC(=O)OC. The atoms list is [C,C,C,C,C,C,C,C,C,C,O,O,C]. If we index the position, the list now is [C1,C2,C3,C4,C5,C6,C7,C8,C9,C10,O11,O12,C13]. The structure matrix is also ordered based on this atoms list with the $i$th row and column being $i$th atom in the atoms list. We can find that for the neighboring atoms in the atoms list, there is always a link. This is also how SMILES is constructed. Since the input data of edges is in the order of C first then O. We change the atoms list of SMILES odering from [C1,C2,C3,C4,C5,C6,C7,C8,C9,C10,O11,O12,C13] to [C1,C2,C3,C4,C5,C6,C7,C8,C9,C10,C11,O1,O2]. Even though the SMILES ordering and mass based ordering are two ordering, they are very similar in our experiment due to the reason that we currently focus on atom C, H, and O. The number of C is usually more than O, so usually the previous several atoms will not change.

With this ordering of the edges in the training label structure matrix ground truth, the network can be guided to learn in a more systematic learning direction. We also tried Singular-Value Decomposition of the label structure matrix and rank the atoms based on corresponding eigenvalues as an alternative ordering for SMILES but did not get good result, so we will omit the details on it here.

\subsection{Other details}
\begin{itemize}
    \item Due to the large number of hydrogens per molecule which makes the whole graph very computational and memory expensive, we will omit them in the SMILES ordering and the graph design as well. Instead, we will add one fake hydrogen atom for each molecule. For example, if the first Carbon is connected to 3 different hydrogen atoms, we will replace it with the Fake Hydrogen Atom connecting to this Carbon with a bond value of three. 
    \item For each atom nodes and mass spectrum nodes, there are multiple messages coming in so we combine the messages through a summation followed by MLP layer, an approximation similar to \cite{dupty2020neuralizing}. For each edge node, number of messages are limited and we can use a multiplication of messages as the original LBP formula.
    \item We add the previous hidden state with the new message as residual connection to improve network learning.
\end{itemize}

\subsection{Overall Architecture}
The architecture of the module is depicted in Figure \ref{factor_graph_neuralize}. Note that initial state of the nodes are initialized through a Message Passing Neural Network as well.

\section{Experiments}

\subsection{Experiment Settings}
The dataset is from the NIST database that contains mass spectrum files and molecular structure information. In our experiment, we focus on the ester functional group with only C H and O atoms. Since we know the molecule is in ester group, we know the substructure of C(=O)O, rendering roughly 1800 molecules in our dataset. So we will extract this substructure and put C(=O)O in the first three position in atoms list.

\subsection{Results}

- \underline{Evaluation}: For single output, the loss may be 0 and 1 for a classification task, with accuracy or recall as evaluation. Since we have multiple outputs for a structure matrix instead of one output, we will use the accuracy to evaluate performance of a single molecule, and average the accuracy of all the molecules  to evaluate the model. For the triangle matrix of 13 atoms, the average accuracy of all-0 output is around 0.84. For loss function we are using is cross-entropy, and we will find minimum loss among the permutation of predicted output matrix.

We use Molecule Attention Structure Elucidation (MASE) as first baseline, an adoption of Molecule Attention Transformer (MAT) \cite{maziarka2020molecule} in which we also include the Mass Spectrum Node and Edge Node in. MASE/MAT has no recurrence or convolution operator, so we need to embed the relative or absolute position of the atoms inside nodes as well. We will vary MASE with options of having positional encoding and decoder or not. Besides, we also use Molecule- Message-Passing-Neural-Network (MMPNN) \cite{gilmer2017neural} as baselines to compare with our FGMN-{Low, Medium, High} Network, as illustrated in Table 1.

\textbf{Note}: FGMN-Low means that the conditional parameter sharing for type B and type C is low, likewise for FGMN-Medium and FGMN-High. Also, SMILES ordering is used instead of Singular-Value-Decomposition to order the atoms since it is observed to boost better performance.

The result is depicted in Table 1, and our FGMN-Medium is the best performance model, where accuracy beats the second-best MASE model by a margin of 2.6\%

% \begin{center}

% \end{center}

\subsection{Ablation studies/ Analysis}

\subsubsection{Valence Factor Type A Effect}
We will analyze the effect of our valence factor type A. We first design a synthetic experimental settings: We will add random exponential distribution noise (represented by beta value) to the ground truth one-hot vector of each edge node, then try if the inference algorithm can decode the result. It is analogous to experiments to show to what extent of noise can Error-Correcting code Inference algorithm can reconstruct the original values.

Table 2 depicts the noise level and the percentage of atoms that have valence constraint perfectly satisfied. Each atom node will have messages coming from multiple factor type A, so Sum Factor refers to messages being summed up while Multiply Factor refers to messages being multiplied. We can see that our decoding algorithm is eliminating the noise to some extent and boosting the performance.

By removing the valence factor type A node from our FGMN-Medium model, the accuracy and percentage of atoms satisfying valence constraints also reduce from 0.914 and 0.387 to 0.892 and 0.295 respectively.

    % \item Removing the factor type B and C.
    % \item Different parameter conditioning for type B and C.

\subsubsection{Factor Type B and C effect}

By training MPNN directly with Valence Factor without factor type B and C, we obtain a testing accuracy as 0.862, with (1.7\%, 5.7\%, 1.0\%) performance margin reduction for FGMN-{Low, Medium, High} models respectively. Therefore, these factors are important since through iterations of message update, it will flow the corresponding information from each mass spectrum peak to the atom nodes of certain substructures, and subsequently flowing to the edges nodes corresponding to these atoms.

\subsubsection{Parameter Sharing Conditioning} 
As explained in \cite{dupty2020neuralizing}, designing good parameter sharing among factors is important. If set of shareable parameter weights is too low, model will lose approximation power to fit the training dataset. Otherwise, if set of shareable parameter weights is too high, we may overfit the training set.

In our experiment (Table 1), FGMN-Medium beats FGMN-Low and FGMN-High by a margin of 4.2\% and 4.9\% respectively.

\section{Conclusion \& Future Work}
In this paper, we have proposed a new architecture that uses various high-order relationships and results in a good network for learning the molecule structure, beating other baselines. Our model also removes the need for using third-party software to get candidate substructures to rank and combine, unlike existing works. 

% We can implement a learnable Permutation Loss function that will permute all possible configurations of label structure matrix and make the process learnable, so removing the need for ordering label structure matrix for asymmetric problem discussed earlier.  
There are various aspects that can be explored to improve this work. For example, we know that permutation of the result structure matrix is a problem, so we can explore ways to embed and learn this knowledge during the training process as well. We might also explore using Reinforcement Learning to sequentially label the edges, further improving the enforcement of the valence constraint.

\bibliography{main}

\begin{thebibliography}{14}
\providecommand{\natexlab}[1]{#1}
\providecommand{\url}[1]{\texttt{#1}}
\providecommand{\urlprefix}{URL }
\expandafter\ifx\csname urlstyle\endcsname\relax
  \providecommand{\doi}[1]{doi:\discretionary{}{}{}#1}\else
  \providecommand{\doi}{doi:\discretionary{}{}{}\begingroup
  \urlstyle{rm}\Url}\fi

\bibitem[{Battaglia et~al.(2018)Battaglia, Hamrick, Bapst, Sanchez-Gonzalez,
  Zambaldi, Malinowski, Tacchetti, Raposo, Santoro, Faulkner
  et~al.}]{battaglia2018relational}
Battaglia, P.~W.; Hamrick, J.~B.; Bapst, V.; Sanchez-Gonzalez, A.; Zambaldi,
  V.; Malinowski, M.; Tacchetti, A.; Raposo, D.; Santoro, A.; Faulkner, R.;
  et~al. 2018.
\newblock Relational inductive biases, deep learning, and graph networks.
\newblock \emph{arXiv preprint arXiv:1806.01261} .

\bibitem[{Dupty and Lee(2020)}]{dupty2020neuralizing}
Dupty, M.~H.; and Lee, W.~S. 2020.
\newblock Neuralizing Efficient Higher-order Belief Propagation.

\bibitem[{Gilmer et~al.(2017)Gilmer, Schoenholz, Riley, Vinyals, and
  Dahl}]{gilmer2017neural}
Gilmer, J.; Schoenholz, S.~S.; Riley, P.~F.; Vinyals, O.; and Dahl, G.~E. 2017.
\newblock Neural message passing for quantum chemistry.
\newblock \emph{arXiv preprint arXiv:1704.01212} .

\bibitem[{Kerber et~al.(2001)Kerber, Laue, Meringer, and
  Varmuza}]{kerber2001molgen}
Kerber, A.; Laue, R.; Meringer, M.; and Varmuza, K. 2001.
\newblock MOLGEN-MS: Evaluation of low resolution electron impact mass spectra
  with MS classification and exhaustive structure generation.
\newblock \emph{Adv Mass Spectrom} 15(939-940): 22.

\bibitem[{Kitson, Larsen, and McEwen(1996)}]{kitson1996gas}
Kitson, F.~G.; Larsen, B.~S.; and McEwen, C.~N. 1996.
\newblock \emph{Gas chromatography and mass spectrometry: a practical guide}.
\newblock Academic Press.

\bibitem[{Krizhevsky, Sutskever, and Hinton(2012)}]{krizhevsky2012imagenet}
Krizhevsky, A.; Sutskever, I.; and Hinton, G.~E. 2012.
\newblock Imagenet classification with deep convolutional neural networks.
\newblock In \emph{Advances in neural information processing systems},
  1097--1105.

\bibitem[{Lim et~al.(2018)Lim, Wong, Wong, Tan, Chieu, Choo, and
  Neo}]{lim2018chemical}
Lim, J.; Wong, J.; Wong, M.~X.; Tan, L. H.~E.; Chieu, H.~L.; Choo, D.; and Neo,
  N. K.~N. 2018.
\newblock Chemical structure elucidation from mass spectrometry by matching
  substructures.
\newblock \emph{arXiv preprint arXiv:1811.07886} .

\bibitem[{MacKay(1999)}]{mncN}
MacKay, D. J.~C. 1999.
\newblock Good Error Correcting Codes based on Very Sparse Matrices.
\newblock \emph{IEEE Transactions on Information Theory} 45(2): 399--431.
\newblock ISSN 0018-9448.

\bibitem[{Maziarka et~al.(2020)Maziarka, Danel, Mucha, Rataj, Tabor, and
  Jastrz{\k{e}}bski}]{maziarka2020molecule}
Maziarka, {\L}.; Danel, T.; Mucha, S.; Rataj, K.; Tabor, J.; and
  Jastrz{\k{e}}bski, S. 2020.
\newblock Molecule Attention Transformer.
\newblock \emph{arXiv preprint arXiv:2002.08264} .

\bibitem[{Pearl(2014)}]{pearl2014probabilistic}
Pearl, J. 2014.
\newblock \emph{Probabilistic reasoning in intelligent systems: networks of
  plausible inference}.
\newblock Elsevier.

\bibitem[{Schymanski et~al.(2008)Schymanski, Meinert, Meringer, and
  Brack}]{schymanski2008use}
Schymanski, E.~L.; Meinert, C.; Meringer, M.; and Brack, W. 2008.
\newblock The use of MS classifiers and structure generation to assist in the
  identification of unknowns in effect-directed analysis.
\newblock \emph{Analytica chimica acta} 615(2): 136--147.

\bibitem[{Schymanski, Meringer, and Brack(2011)}]{schymanski2011automated}
Schymanski, E.~L.; Meringer, M.; and Brack, W. 2011.
\newblock Automated strategies to identify compounds on the basis of GC/EI-MS
  and calculated properties.
\newblock \emph{Analytical Chemistry} 83(3): 903--912.

\bibitem[{Varmuza and Werther(1996)}]{varmuza1996mass}
Varmuza, K.; and Werther, W. 1996.
\newblock Mass spectral classifiers for supporting systematic structure
  elucidation.
\newblock \emph{Journal of Chemical Information and Computer Sciences} 36(2):
  323--333.

\bibitem[{Vaswani et~al.(2017)Vaswani, Shazeer, Parmar, Uszkoreit, Jones,
  Gomez, Kaiser, and Polosukhin}]{vaswani2017attention}
Vaswani, A.; Shazeer, N.; Parmar, N.; Uszkoreit, J.; Jones, L.; Gomez, A.~N.;
  Kaiser, {\L}.; and Polosukhin, I. 2017.
\newblock Attention is all you need.
\newblock In \emph{Advances in neural information processing systems},
  5998--6008.

\end{thebibliography}
\end{document}